\documentclass{article}

\usepackage{arxiv}

\usepackage[utf8]{inputenc} 
\usepackage[T1]{fontenc}    
\usepackage{hyperref}       
\usepackage{url}            
\usepackage{booktabs}       
\usepackage{amsfonts}       
\usepackage{nicefrac}       
\usepackage{microtype}      
\usepackage{lipsum}		
\usepackage{graphicx}
\usepackage{natbib}
\usepackage{doi}
\usepackage{algorithm}
\usepackage{algpseudocode}
\usepackage{amsmath}

\title{Multi-UAV Conflict Resolution with Graph Convolutional Reinforcement Learning}

\author{ \href{https://orcid.org/0000-0003-0839-3235}{\includegraphics[scale=0.06]{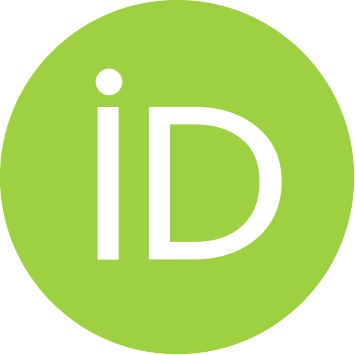}\hspace{1mm}Ralvi Isufaj}\thanks{Corresponding author.} \\
	Logistic and Aeronautics Group\\
	Autonomous University of Barcelona\\
	Sabadell, Spain \\
	\texttt{ralvi.isufaj@uab.cat} \\
	\And
	\href{https://orcid.org/0000-0002-4547-3202}{\includegraphics[scale=0.06]{orcid.pdf}\hspace{1mm}Marsel Omeri} \\
	Logistic and Aeronautics Group\\
	Autonomous University of Barcelona\\
	Sabadell, Spain \\
	\texttt{marsel.omeri@uab.cat} \\
	\And
	\href{https://orcid.org/0000-0002-7227-7944}{\includegraphics[scale=0.06]{orcid.pdf}\hspace{1mm}Miquel Angel Piera} \\
	Logistic and Aeronautics Group\\
	Autonomous University of Barcelona\\
	Sabadell, Spain \\
	\texttt{miquelangel.piera@uab.cat} \\
}

\renewcommand{\shorttitle}{Multi-UAV Conflict Resolution with Graph Convolutional Reinforcement Learning}

\hypersetup{
pdftitle={A template for the arxiv style},
pdfsubject={q-bio.NC, q-bio.QM},
pdfauthor={David S.~Hippocampus, Elias D.~Striatum},
pdfkeywords={First keyword, Second keyword, More},
}

\begin{document}
\maketitle

\begin{abstract}
	Safety is the primary concern when it comes to air traffic. In-flight safety between Unmanned Aircraft Vehicles (UAVs) is ensured through pairwise separation minima, utilizing conflict detection and resolution methods. Existing methods mainly deal with pairwise conflicts, however due to an expected increase in traffic density, encounters with more than two UAVs are likely to happen. In this paper, we model multi-UAV conflict resolution as a multi-agent reinforcement learning problem. We implement an algorithm based on graph neural networks where cooperative agents can communicate to jointly generate resolution maneuvers. The model is evaluated in scenarios with 3 and 4 present agents. Results show that agents are able to successfully solve the multi-UAV conflicts through a cooperative strategy.
\end{abstract}


\section{\label{sec:intro} Introduction}
Commercial and civil unmanned aircraft systems (UAS) applications are projected to have significant growth in the global market. According to SESAR, the European drone market will exceed \texteuro10 billion annually by 2035, and over \texteuro15 billion annually by 2050 \citep{SESARJU2016}. Furthermore, considering the characteristics of the missions and application fields, it is expected that the most market value will be in operations of small UAS (sUAS) and at the very-low-level airspace (VLL). Such a growing trend will be accompanied by an increase in traffic density and new challenges related to safety, reliability, efficiency. Therefore, the development and implementation of conflict management systems are considered a pre-condition to integrate UAS in the civil airspace.
Most notably, the National Aeronautics and Space Administration (NASA) in the USA aims to create a UAS Traffic Management (UTM) system that will make it possible for many UAS to fly at low altitudes along with other airspace users\citep{Barrado2020U-spaceOperations}. Europe is leading efforts to develop an equivalent UTM concept, referred to as U-space. It will provide a set of services (and micro-services) that would accommodate current and future traffic (mainly but not limited to) at VLL airspace\citep{Prevot2016UASOperations}. Similar approaches are followed also in China and Japan \citep{Zhang2018UOMSChina}. 
Considering airspace under UTM services, UAS must be capable of avoiding static conflicts such as buildings, terrain, and no-fly zones and dynamic conflicts such as manned or unmanned aircraft. Here a pairwise conflict is defined as a violation of the en-route separation minima between two UAVs \citep{icao}. To ensure operations free of conflict, UTM provides Conflict Detection and Resolution services, which comprise three layers of safety depending on the time-horizon (i.e.look-ahead time) \citep{nasa}: Strategic and Tactical Conflict Mitigation and Collision Avoidance (CA)\citep{icao}\citep{nasa}. 
In this work, we will focus on tactical CR applicable for small UAS missions. This function is typically treated in two ways: Self-separation and Collision Avoidance\citep{nasa}\citep{Radanovic_2019}.  The former is a maneuver executed seconds before the loss of separation minima, characterized by a slight deviation from the initial flight plan, and aims to prevent CA activation. The latter provides a last-resort safety layer characterized by imminent and sharp escape maneuvers. Both functions above are encompassed within what is widely recognized as Detect and Avoid capability \citep{7778033, Johnson2017Johnson_DetectAndAvoidHandout}.
Aligning with the up-to-date state-of-the-art, a loss of separation minima is referred to as loss of Well Clear (LoWC). While there is no standard definition of well clear (WC), two related functions are associated with this state: Remain Well Clear (RWC), and Collision Avoidance (CA)\citep{Manfredi2017}. In terms of tactical CD\&R, RWC is equivalent to the self-separation function.
Defining and computation of RWC thresholds is an open research work but is mainly viewed as protection volume around UAS\citep{Cook2017,Consiglio,Muoz2016UnmannedAS}. This volume can be specified by spatial thresholds, temporal thresholds, or both at the same time. We follow the hockey-puck model\citep{Weinert2018Well-clearRisk,McLain2017AWC} characterized by distance-based thresholds. In addition, the near-mid-air-collision (NMAC) represents the last safety volume. As the name suggests, a distance smaller than NMAC represents a very severe loss of well clear that could result in a collision in the worst case. This distance is usually defined based on the dimensions of the UAS and its navigation performance\citep{Modi2016ApplyingRN}.

There are many existing works that propose conflict resolution algorithms (see Section II for a more detailed overview). However, the majority of these works focus mainly on pairwise conflicts. Nevertheless, with the expected increase in traffic density \citep{SESARJU2016} multi-UAV conflicts (i.e. involving more than 2 UAVs) are expected to occur. In this paper, multi-UAV conflict resolution is modeled as a multi-agent reinforcement learning problem (MARL). More specifically, we utilize graph convolutional reinforcement learning \citep{jiang2018graph}, where air traffic is modeled as a graph. The present UAV are the set of nodes, and single pairwise conflicts form the set of edges in the graph. The model used in this paper provides a communication mechanism between connected nodes in the graph. Such a mechanism facilitates learning and allows for the agents\footnote{In this work, an agent is an abstraction of a UAV. An agent must learn how to achieve a policy that solves conflicts through training in several multi-UAV conflict scenarios. The generated maneuvers are then forwarded to the UAVs in the actual scenario} to develop cooperative strategies. Multi-UAV conflicts are formally defined as compound conflicts, where multiple pairwise conflicts have tight spatial and temporal boundaries. In this work, we first train a model in scenarios with three UAVs. After that, the same model is retrained to solve compound conflicts with four UAVs. This technique allows us to re-use the previously learned policies and refine them to a new set of scenarios while efficiently training the new agent from scratch. Results show that agents are successfully able to solve compound conflicts in both cases.

The rest of the paper is organized as follows: some existing works are discussed in Section 2. Section 3 describes the theoretical background necessary for this paper. In Section 4, the experimental setup is presented. Results are presented and discussed in section 5, while in Section 6, we draw conclusions and propose steps for further research.

\section{Related Work}
There are many essential contributions in the area of conflict resolution methods in aviation. These methods are widely classified into the geometric, force field methods, optimized trajectory, and Markov Decision Process (MDP) approaches (probabilistic) \citep{skowron2019sense}. For detailed and comprehensive information on CD\&R practices, we suggest Kuchar and Yang’s review study \citep{kuchar2000review} and this review paper \citep{ribeiro2020review} for more up-to-date content. We will focus only on the MDP method and provide a summary discussion below, as our work aligns with this group of methods.  
Aircraft and especially UAS operations are characterized by uncertain environments and stochastic events such as weather, multiple intruders, Communication, Navigation, and Surveillance (CNS) failures; therefore, decision making methods that adapt under such conditions are necessary. MDP and more recent Partial Observable MDP (POMDP) are methods that can have significant performance in such domains. Different techniques are used to solve MDP and/or POMDP problems, and most noticed are reinforcement learning (RL) and deep reinforcement learning (DRL) methods. In \citep{bertram2020distributed}, the authors present an efficient MDP-based algorithm that provides self-separation functions for UAS in free airspace. A similar approach is followed here \citep{yang2020scalable}, where the authors give a scalable multi-agent computational guidance for separation assurance in Urban Air Mobility. In addition, they use RL techniques to solve the MDP problems. In a previous work \citep{hu2020uas}, a conflict resolution system is applied to mitigate conflicts between UAS.  Ribeiro et al. \citep{ribeiro2020determining} consider a single-agent approach to conflict resolution through RL for unmanned aerial vehicles (UAVs). Furthermore, recent works have seen the engagement of DRL methods, which behave better in multi-agent environments and consider uncertainties. In this paper \citep{isufaj2021towards}, the authors model pairwise conflict resolutions as a multi-agent reinforcement learning (MARL) problem. They use Multi-Agent Deep Deterministic Policy Gradient (MADDPG) \citep{lowe2017multi} to train two agents, representing each aircraft in a conflict pair, capable of efficiently solving conflicts in the presence of surrounding traffic by considering heading and speed changes. In \citep{pham2019machine}, the authors use the Deep Deterministic Policy Gradient (DDPG) technique to mitigate conflicts in high density scenarios and uncertainties. Brittain et al. \citep{brittain2020deep} used a deep multi-agent reinforcement learning framework to ensure autonomous separation between aircraft. Dalmau et al. \citep{dalmauair} used Message Passing Neural Networks (MPNN) to model air traffic control as a multi agent reinforcement learning system where agents must ensure conflict free flight through a sector. 

While these papers consider a multi-aircraft (manned or unmanned) setting, they do not particularly consider small UAS performance capabilities (i.e., high yaw rate). Also, a common assumption is that the flight trajectories should be within a predefined airspace sector. In a UTM environment airspace is not necessarily segregated into sectors. Additionally, small UAS characteristics can directly effect how the action space is modelled. Moreover, approaches with a multi-UAV setting do not consider the effects of cooperation on the resolution manoeuvres. In this work, we propose a multi-UAV conflict resolution method suitable for sUAS operations and attempt to achieve cooperation between the agents. 

It is worth noting that these methods (RL and DRL) are considered very important for the development of the Aircraft Collision Avoidance System (ACAS-X), which will be extended into ACAS-Xu, ACAS-sXu, and so on, to accommodate all airspace users \citep{manfredi2016introduction}.

\section{Theoretical Background}
\subsection{Reinforcement Learning}
Reinforcement Learning (RL) is a paradigm of machine learning which deals with sequential decision making \citep{sutton1998introduction}. A given RL problem is formalized by a Markov Decision Process (MDP), which is a discrete time stochastic control process \citep{bellman1958dynamic} that consists of a 4-tuple $(S, A, T, R)$, where:
\begin{itemize}
    \item $S$ is the state space,
    \item $A$ is the action space,
    \item $T: S \times A \times S \rightarrow [0,1]$ is the transition function which is a set of conditional probabilities between states,
    \item $R: S \times A \times S \rightarrow \mathbb{R}$ is the reward function
\end{itemize}
In RL, an agent makes decisions in an environment to maximize a certain notion of cumulative reward G, defined as follows:
\begin{equation}
    G_t = R_{t+1} + \gamma R_{t+2} + \gamma^2 R_{t+3} + ... = \sum_{k=0}^{\infty} \gamma^k R_{t+k+1}
\end{equation}
where $\gamma$ is a discount factor between 0 and 1. Its task is to inform the agent how relevant immediate rewards are in relation to rewards further in the future. The higher $\gamma$ is the more the agent will care about future consequences.

The agent improves incrementally by modifying its behaviour according to previous experience. The agent does not strictly require complete information or knowledge of the environment, it only needs to interact with it and gather information \citep{franccois2018introduction}. 

The RL agents starts at an initial state $s_0 \in S$ and at each time step $t$ must take an action $a_t \in T$. Then, the agent gets a reward $r_t \in R$ from the environment. The states then transitions to $s_{t+1} \in S$, which is dictated by the taken action and the dynamics of the environment. Finally, the agent stops interacting with the environment when it reaches a defined goal state.

The agent's behavior is encoded into a policy $\pi$, which can be deterministic $\pi: S \rightarrow A$, or stochastic $\pi: S \times A \rightarrow [0,1]$. 

There are two ways that are used to predict the total future discounted reward: the value function $V^{\pi}$ and the action-value function $Q^{\pi}$, defined as follows:
\begin{equation}
    V^{\pi}(s) = \mathbb{E}_{\pi}(R_t \vert s_t=s)
\end{equation}
\begin{equation}
    Q^{\pi}(s,a) = \mathbb{E}_{\pi}(R_t \vert s_t=s, a_t=a)
\end{equation}
The value function represents the future expected reward in the current state if the policy $\pi$ is followed, while action-value function represents expected rewards for state-action pairs following policy $\pi$. Ultimately, the goal of all RL algorithms is to solve either of these functions.

Q-learning is one most prominent algorithms for solving RL problems. There, an agent must learn to estimate the optimal action-value function in the form of a table with as many state-action pair entries as possible \citep{dalmauair}. However, in cases where the state space or action space (or both) are continuous, there are infinitely many state-action pairs, which makes it unfeasible to store the values in table. In those cases, a function is used to approximate the Q function. Such a function with parameters $\mu$, is optimized through an objective function based on the Bellman equation \citep{bellman1958dynamic}. 

In the case of Deep Q-Networks (DQN) \citep{mnih2013playing}, the Q function approximators are neural networks. However, several issue arise when applying deep learning directly on a RL problem. First, in RL rewards can be sparse or delayed, which hinders neural networks, as they rely on directly gained feedback. Additionally, the data that is obtained from a RL problem is highly correlated and lastly, the data distribution changes as the policy does, making it non-stationary which further impairs the learning capabilities of neural networks. In order to overcome these issues, several modifications must be. Experience replay is used to mitigate the issue of sample autocorrelation \citep{mnih2013playing}. In this technique, the agent's experience is stored at each time step in a replay buffer. the memory is sampled randomly and is used to update the networks. When the replay buffer becomes full, the simplest solution is to discard the oldest samples. The non-stationarity of the data makes the training unstable, which can lead to undesired phenomena such as \textit{catastrophic forgetting}, where the agent suddenly "forgets" how to solve the task after apparently having learned a suitable policy. Such an issue can be mitigated using \textbf{target networks}, which is an identical network to the one used to learn the Q function, that is held constant to serve as a stable target for learning for a fixed number of time steps.
\subsection{Multi-Agent Reinforcement Learning}
Multi-Agent Reinforcement Learning (MARL) is an extension of classical RL where there are more than one agents in the environment. This is formalized through partially observable Markov games \citep{littman1994markov}, which are decision processes for $\mathbb{N}$ agents.

Similarly to MDPs, Markov games have a set of actions. However, in this case, the environment is not fully observable by the agents. Therefore, the Markov game has a set of observations $O_1,...,O_N$ for each agent. Similarly to single agent RL, in the MARL setting agents take actions according to their policy and obtain rewards. The goal of the agents is to maximize personal and total expected reward.
\subsection{Graph Convolutional Reinforcement Learning}
While deep learning has proven effective in capturing patterns of Euclidean data, there are a number of applications where data are represented as graphs \citep{wu2020comprehensive}. The complexity of graph data has imposed significant challenges on existing deep learning algorithms. A graph can be irregular and dynamic, as it can have a variable number of nodes and the connections between nodes can change over time. Furthermore, existing deep learning algorithms largely assume the data to be independent, which does not hold for graph data.

Recently, there has been an increasing number of works that extend deep learning approaches to graph data, called Graph Neural Networks (GNNs). Variants include: Graph Attention Networks (GATs) \citep{velivckovic2017graph}, Graph Convolutional Networks (GCNs) \citep{kipf2016semi} and Message Passing Neural Networks (MPNNs) \citep{gilmer2017neural}. We refer the reader to \citep{wu2020comprehensive}, for a comprehensive review of GNNs.

In the case of MARL, communication is often cited as a key ability for cooperative agents \citep{jiang2018graph, dalmauair}. In such a setting, agents exchange information before taking an action. 

In this work, we will use Graph Convolutional Reinforcement Learning \citep{jiang2018graph} (dubbed DGN by its authors), which is a GNN algorithm for cooperative agents. 

In DGN, the multi-agent environment is modeled as a graph $\mathcal{G} = (\mathcal{V},\mathcal{E})$, where $\mathcal{V}$ is the set of nodes and $\mathcal{E}$ is the set of edges. Each agent is a node and the local observation of the agent are the features of the node. Each node $i$ has a set of neighbors $\mathbb{B}_i$, where
$(i,j) \in \mathcal{E}$, $\forall j \in \mathbb{B}_i$. The set of neighbors is defined according to some criteria, depending on the environment and changes over time. In DGN, neighbor nodes can communicate with each other. Such a choice leads to the agents only considering local information when making their decisions. Another option would be to consider all agents in the environment, however, this comes with higher computational complexity.

DGN has three modules: an observation encoder, convolutional layer and Q network. The observation of an agent $i$ at time step $t$, $o_i^t$ is encoded into a feature vector $h_i^t$ by a Multi Layer Perceptron (MLP). The convolutional layer combines the feature vectors in the local region and generates a latent feature vector $h'_{i^t}$. The receptive field of the agents increase by stacking more convolutional layers on top of each other. An important property of the convolutional layer is that it should be invariant from the order of the input feature vectors. Furthermore, such a layer must be effective in learning how to abstract the relation between agents as to combine the input features.

DGN uses multi-head dot-product attention \citep{zambaldi2018relational}, which is an implementation of attention which runs the attention mechanism several times in parallel, to compute interactions between agents (we refer the reader to \citep{jiang2018graph, zambaldi2018relational} for a detailed overview of the attention mechanism). Let us denote with $\mathbb{B}_{+i}$ the set of neighbors $\mathbb{B}_i$ and agent $i$. The input features of the agent $i$ are projected into query $Q$, key $K$  and value $V$ representation by every attention head. For an attention head $m$ the relation for $i,j \in \mathbb{B}_{+i}$ is as follows:
\begin{equation}
    \alpha_{ij}^m = \frac{exp(\tau \cdot W_Q^mh_i \cdot (W_K^mh_j)^T)}{\sum_{a \in \mathbb{B}_{+i}}  exp(\tau \cdot W_Q^mh_i \cdot (W_K^mh_a)^T)))}
\end{equation}
where $\tau$ is a scaling factor and $W_Q^m$ and $W_K^m$ are the weight matrices of the query and key for attention head $m$. The representations of the input features are weighted by the relation and summed together, which is done for each head $m$. The outputs of all attention heads for an agent $i$ are concatenated and then fed into a MLP $\sigma$ as follows:
\begin{equation}
    h'_i = \sigma(concatenate (\sum_{a \in \mathbb{B}_{+i}} \alpha_{ij}^m W_{V}^m h_a, \forall m \in M))
\end{equation}
The graph representing the agents and the interactions between them is formalized through and adjacency matrix $C$, where the $i$th row contains a 1 for each agent in $\mathbb{B}_i$ and 0 for any agents not in the neighborhood of $i$. The feature vectors are merged into a feature matrix $F$ with size $N \times L$ where $N$ is the number of agents and $L$ is the length of the feature vector. The feature vectors in the local region of agent $i$ are obtained by $C_i \times F$.

The Q network in DGN is a common network as described in II.B. However, in DGN, the outputs of the graph convolution layer are concatenated and fed into the network. At each time step, the tuple $(O, A, O', R, C)$ is stored in the replay buffer, where $O$ and $O'$ are the current and next observations, $A$ is the set of actions, $R$ isthe set of rewards and $C$ is the adjacency matrix. During training, a random minibatch of size S is sampled from the buffer and the loss is minimized as follows:
\begin{equation}
    \mathcal{L}(\theta) = \frac{1}{S} \sum_S \frac{1}{N} \sum_{i=1}^N (y_i - Q(O_{i,C,a_i};\theta))^2
\end{equation}
where $y_i$ indicates the return. Another factor that can impact the training of the Q network is the dynamic nature of the graph, which can change from one time step to the other. To mitigate this, the adjacency matrix $\mathcal(C)$ is kept unchanged in two successive time steps when computing the Q values in training. Finally, the target network with parameters $\theta'$ is updated from the Q network with parameters $\theta$ as follows:
\begin{equation}
    \theta' = \beta \theta + (1 - \beta)\theta'
\end{equation}
where $\beta$ indicates the importance of the new parameters in the target network. 
\section{Experimental Setup}
\subsection{Compound Conflicts}
In this work, we consider multi-UAV conflicts. However, multiple pairwise conflicts can have varying spatial and temporal boundaries, i.e., their overlap in space and time. Koca et al.\citep{kocastrategies}, introduce the concept of a \textit{compound ecosystem}, with an ecosystem being the set of aircraft affected by the occurrence of a conflict. They propose that multiple ecosystems can be considered together if they have at least one common member and the conflicts overlap in time more than 10\% of their duration. 
For this work, we relax the requirements by not considering surrounding traffic, therefore proposing the concept of a \textit{compound conflict}. As such, multiple pairwise conflicts can be considered collectively if and only if they share a common aircraft. We keep the temporal requirement the same as in \citep{kocastrategies}.

\subsection{Traffic as a Graph}
In this work, the multi-agent environment is represented as a graph. Therefore, we must define how the graph is created for a given traffic scenario. In order to have a correct definition of a graph $\mathcal{G}=(\mathcal{V},\mathcal{E})$, the set of nodes and edges is required.

In DGN, the nodes are the agents present in the environment. We keep the same approach by considering the UAVs as nodes in a given traffic scenario. An edge is created between two UAVs if and only if a conflict between them has been detected. This choice is motivated by the fact that in DGN, agents communicate with their neighbors first and foremost. Therefore, we make this choice to facilitate cooperation between UAVs that are in conflict.

\subsection{Training Environment}
In order to model compound conflict resolution as a MARL problem, the underlying Markov decision process must be formalized. Thus, we have to determine the state space, action space, and reward function. As we are considering cooperative agents, the ultimate goal is to maximize the joined reward. Therefore, all agents have the same reward structure.
\subsubsection{State space}
The representation of the states of the environment is one of the most critical factors that can impact the learning capability and performance of the agents. Typically, the state is formalized through a vector of a certain dimensionality which should provide enough information to facilitate learning. Nevertheless, representations with higher dimensionality will suffer from a higher computational effort to train an effective model.

Therefore, in this work we take the state representation proposed by Isufaj et al.\citep{isufaj2021towards}, where the state is formalized through the agents' position and speed information. More specifically the state $s_i$ of an agent is the vector $s_i=[lat,lon,hdg,spd]$, i.e. latitude, longitude, heading and speed. These values are normalized into the range $[0,1]$ to make it easier for the model to be trained.
\subsubsection{Action space}
In this work, we only consider solutions through heading changes, thus speed and altitude changes are ignored. As such, agents can choose to take on of three actions at each decision time step: turn left, turn right, do nothing, where each track change corresponds to a heading change of $15^o$ in either direction. Agents must make a decision every 2 seconds. 
\subsubsection{Reward function}
Once the agents take an action according to their policy, they will receive a reward from the environment $r_{i,t}$ for the current time step, which indicates the quality of the action. Thus, a carefully constructed reward function is crucial in achieving desirable performance \citep{isufaj2021towards}.

In our case, the reward consists of three terms. First of all, the \textit{number of conflicts} term punishes agents according to the number of conflicts. The more conflicts the agent is in, the more it will have the incentive to solve the conflicts. Furthermore, through the \textit{severity term}, the agents are encouraged to solve the most severe conflicts first. This term considers more severe conflicts, i.e., smaller CPA distance, as more important to solve first. Lastly, the \textit{deviation term} penalizes the agents for solutions that drift the agent from its original track. In this work, if an agent has deviated more than $90^o$ from the original route, it is penalized heavily. In cases where it hasn't, it is penalized as a fraction of the current deviation to the maximal deviation. Such a term indirectly also incentivizes the agents to solve the conflicts as soon as possible, as the quicker the conflicts are solved, the less of a negative reward the agent will get. Formally, the reward function is as follows:
\begin{equation}
\begin{split}
    r_{i,t} = w_1\sum_{\mathcal{E}(i)}-1 + w_2 \begin{cases} -\frac{\vert \mu - \mu' \vert}{90} &\text{if $\vert \mu - \mu'\vert < 90$} \\ -10 &\text{otherwise} \end{cases}  \\+ w_3 (1-exp(1 - \frac{1}{(\frac{d_{cpa}}{d_{thresh}})^{1/2}}))
\end{split}
\end{equation}
where $w_1, w_2, w_3$ are positive weights that indicate the importance of each term, $\mathcal{E}(i)$ indicates all the agents that have an edge with $i$, $\mu$ and $\mu'$ are the original and current heading and $d_{cpa}$ and $d_{thresh}$ are the CPA and self-separation distances. In this paper, $w_1, w_2, w_3$ are kept equal, however in future work these can be extended to be learnable parameters. The total reward for a given time step $t$ is:
\begin{equation}
    R_t = \sum_i^N r_{i,t}
\end{equation}
where $N$ is the total number of agents.
\subsection{Simulation Environment}
Simulations were run on the Air Traffic Simulator BlueSky \citep{hoekstra2016bluesky}. The simulator was chosen primarily because it is an open-source tool, allowing for more transparency in developing and evaluating the proposed model. Furthermore, BlueSky has an Airborne Separation Assurance System (ASAS), supporting different CD\&R methods. This allows for different resolution algorithms to be evaluated under the same conditions and scenarios.
\subsection{Data Generation}
\begin{algorithm}
\caption{Data Generation Algorithm}\label{alg:gen}
\begin{algorithmic}
\Procedure{GENERATE}{$target,spd_{min},spd_{max},t_{loss},t_{la}$}
\State $created\gets \emptyset$
\State $hdgs_{conflict}\gets [0,45,90,135,180,-45,-135,-90]$
\State $var\gets$10
\State $lat_{ref}\gets 41.4$
\State $lon_{ref}\gets 2.15$
\State $spd_{ref}\gets $UNIFORM$(speed_{min},speed_{max})$
\State $hdg_{ref}\gets $UNIFORM$(1,360)$
\State $ac_{ref}\gets$AIRCRAFT$(lat_{ref},lon_{ref},spd_{ref},hdg_{ref})$
\State $created\gets$ $created$ $\cup$ $ac_{ref}$
\While {SIZE($created$) $<$ $target$}
\State $accepted\gets$False
\While{$accepted$ is False}
\State $hdg\gets$SAMPLE($hdgs_{conflict}$)+UNIFORM(-$var$, $var$)
\State $severity\gets$UNIFORM(0.1,1)
\State $cpa\gets threshold - (threshold \times severity)$
\State $chosen\gets$SAMPLE($created$)
\State $ac_{proposed}\gets$CRECONF($hdg,cpa,tloss,chosen$)
\State $in_{conf}\gets$CONFLICT($created,ac_{proposed}, t_{la})$
\If{$in_{conf}$ is False}
\State $accepted\gets$True
\State $created\gets created$ $\cup$ $ac_{proposed}$
\EndIf
\EndWhile
\EndWhile
\State \textbf{return} $created$
\EndProcedure
\end{algorithmic}
\end{algorithm}
Algorithm \ref{alg:gen} describes the procedure to generate the training scenarios. In this work, we consider compound conflicts with 3 and 4 UAVs. To create the multi-UAV conflict, first, a reference aircraft is initialized, with a heading sampled from a uniform distribution from $0^o$ to $360^o$. Then, this aircraft is added to the set of created aircraft. 
To generate the rest of the conflicting UAVs, we sample from the set of the created ones. Then, a conflict angle is chosen from the list $[0^o, 45^o, 90^o, 90^o, 135^o, 180^o, -135^o, -45^o]$. Next, to add some variance to the intrusion headings, a variance in the range $[-10^o, 10^o]$ is added to each case. After that, the severity of the conflict is decided by sampling from a uniform distribution between 0.1 and 1. Finally, we set the time the new aircraft enters in conflict with the randomly chosen aircraft to 15 seconds. The CRECONF function is taken from the BlueSky simulator, and it provides the location and speed of a new conflicting aircraft. However, as compound conflicts have temporal boundaries, no accidental conflicts are added in one look-ahead time, which is set to 8 seconds. This is checked by the CONFLICT function, also taken from BlueSky.
To define the metrics for self-separation, we follow a similar approach as in\citep{inproceedings,Mullins2013DynamicST}. This threshold depends on the UAV maneuverability and its maximum airspeed. Whereas the innermost layer will be modeled according to the Near Mid Air Collision concept, as a circle with radius: $R_{NMAC}$ = 2 $\times$ Maximum Wing Span + Total System Error (TSE). 
The self-separation can be calculated by the equation below: 
\begin{equation}
    R_t = R_{NMAC} + V_{m} \times t_m + \frac{V_m}{\omega_m}
\end{equation}
where $V_m$ and $\omega_m$ are maximum airspeed and maximum yaw rate, respectively. Whereas $t_m$ is the time needed for the UAV to make an avoidance maneuver. 
The self-separation threshold was set to 240m, taking into account $R_{NMAC}$ = 4m, maximum airspeed 15m/s, and a maximum yaw rate of $90^o/s$ As we are attempting to solve conflicts at the tactical level, a duration of 1 minute per scenario was deemed suitable. Note that the time metrics (i.e tactical CD\&R maneuver and look-ahead time) mentioned above are synthesized from the state of the art of CD\&R in small UAS \citep{ICDR,Consiglio,Johnson2017ExplorationOD}.

\section{Simulation Results}
\subsection{Conflict Resolution Performance}
The model was trained for 10000 episodes with scenarios of compound conflicts with 3 UAVs. Then, it was trained for a further 10000 episodes with scenarios of compound conflicts with 4 UAVs. In this way, we utilize the learned policies of the previous agents to fine-tune them in the four-agent case and train the new agent from scratch. The models were trained on the Google Cloud Platform\footnote{https://cloud.google.com} using an NVIDIA Tesla K80 GPU. The training lasted around 10 hours.
\begin{figure}[h]
\centering
\includegraphics[width=\linewidth]{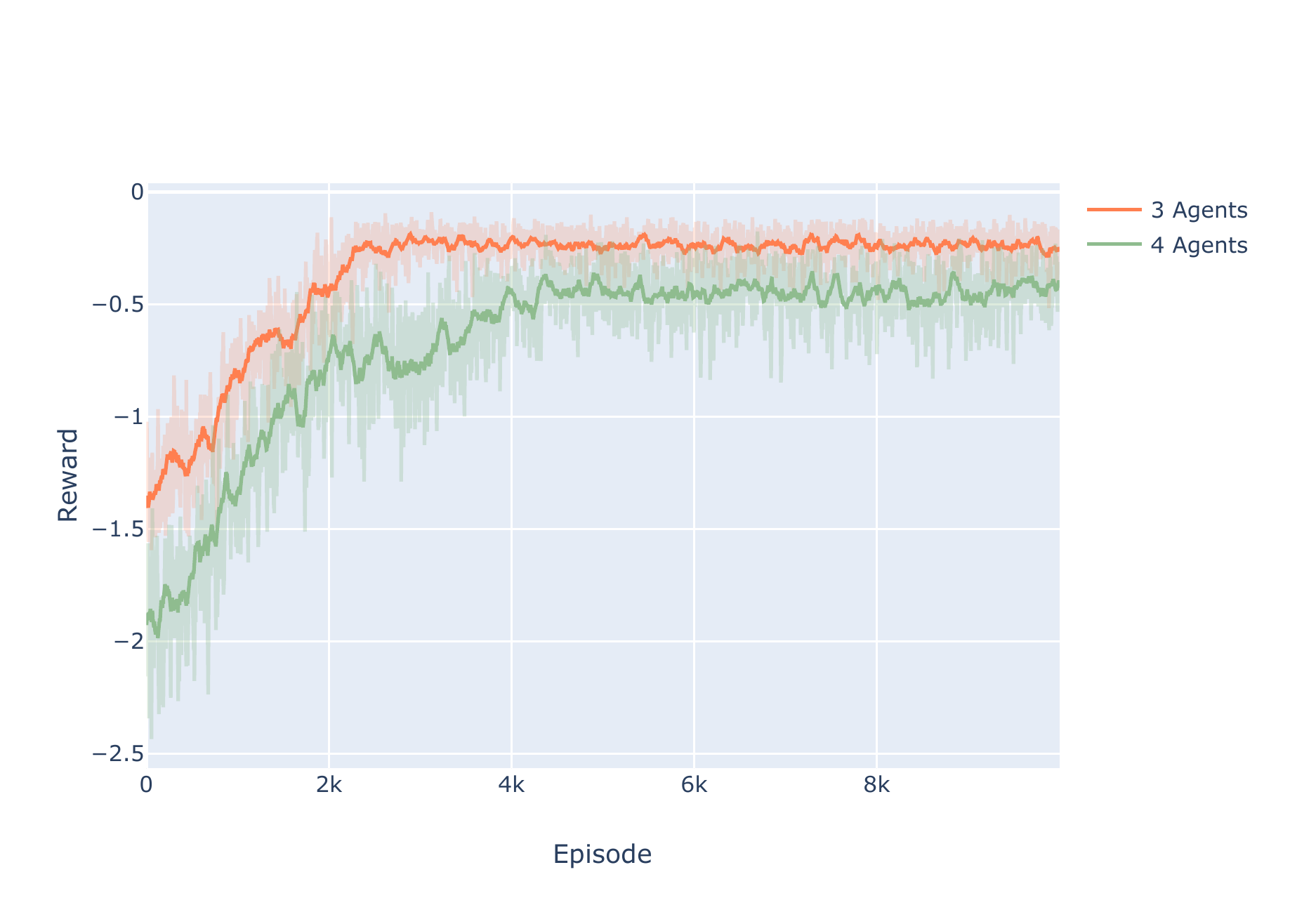}
\caption{Evolution of the cumulative reward per episode.}
\label{fig:reward}
\end{figure}

Figure \ref{fig:reward} shows the evolution of the cumulative reward for both cases. As the agents are cooperative, we are interested in the overall reward that is gained per episode and do not concern ourselves with the individual rewards. In this work, we utilize negative rewards, so the maximum the agents can get is 0. In the case of the 4 agents the reward seems a bit lower, however this comes a result of there being one more agent present, which takes actions to solve the conflicts thus inflicting itself some negative rewards for going away from track.  

From the figure, it can be noted that the model converges on both occasions. This means that the agents are successfully able to improve their policies with gained experience. However, in the case with 3 present agents, the convergence happens around 2000 episodes, while around 4000 episodes are required for the 4 agent case. In the latter case, there are more possible scenarios that can be generated, therefore increasing the variance of situations that the agents are presented with. Furthermore, in the beginning of training the already present agents employ their learned policies, while the new agent is exploring the possible actions, which reduces the overall reward the agents get.

\begin{figure}[h]
\centering
\includegraphics[width=\linewidth]{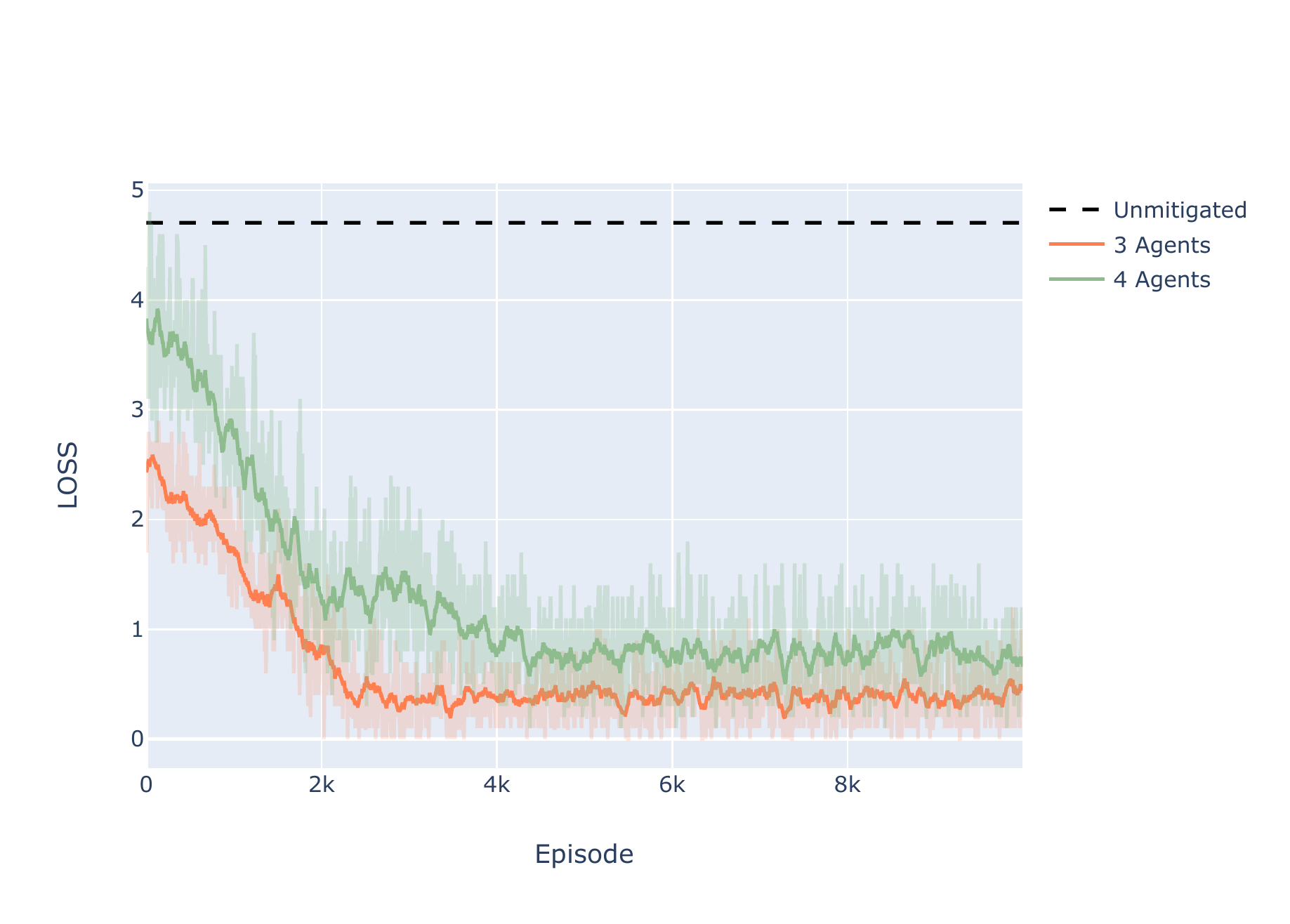}
\caption{Number of losses of separation in comparison with the average unmitigated case.}
\label{fig:conf}
\end{figure}

In Figure \ref{fig:conf} the number of losses of separation (LOSS) is shown. The number of LOSS of the average unmitigated case (for both 3 and 4 agents) is shown with the dashed line. As we can see, the reward performance translates directly to successfully avoiding LOSS. In the case with 3 present agents, after convergence the average LOSS per episode is less than 1. This indicates that the agents are able to successfully solve conflicts before violating the self-separation distance. In the case of the 4 agents compound conflict, the average is around 1 LOSS per episode. However, through our results, we note that the model manages to always avoid near misses in both cases, as the NMAC distance is never breached. 

It is interesting to note that we can observe a similar evolution as in Figure \ref{fig:reward}, with the retrained model performing slightly worse. In general, the case with 4 agents had more pairwise conflicts present, which makes the problem more difficult. 

Table \ref{noloss} shows in how many episodes the compound conflict was solved, meaning no LOSS has occurred. We note that the difference is similar to the number of extra epochs the 4 agents model needed to converge. As such, once the model converges it can generally manage to solve the compound conflict, thus fulfilling its task successfully. This result shows that the agents are able to solve conflicts through communicating with their neighbors. 

\begin{figure}[h]
\centering
\includegraphics[width=\linewidth]{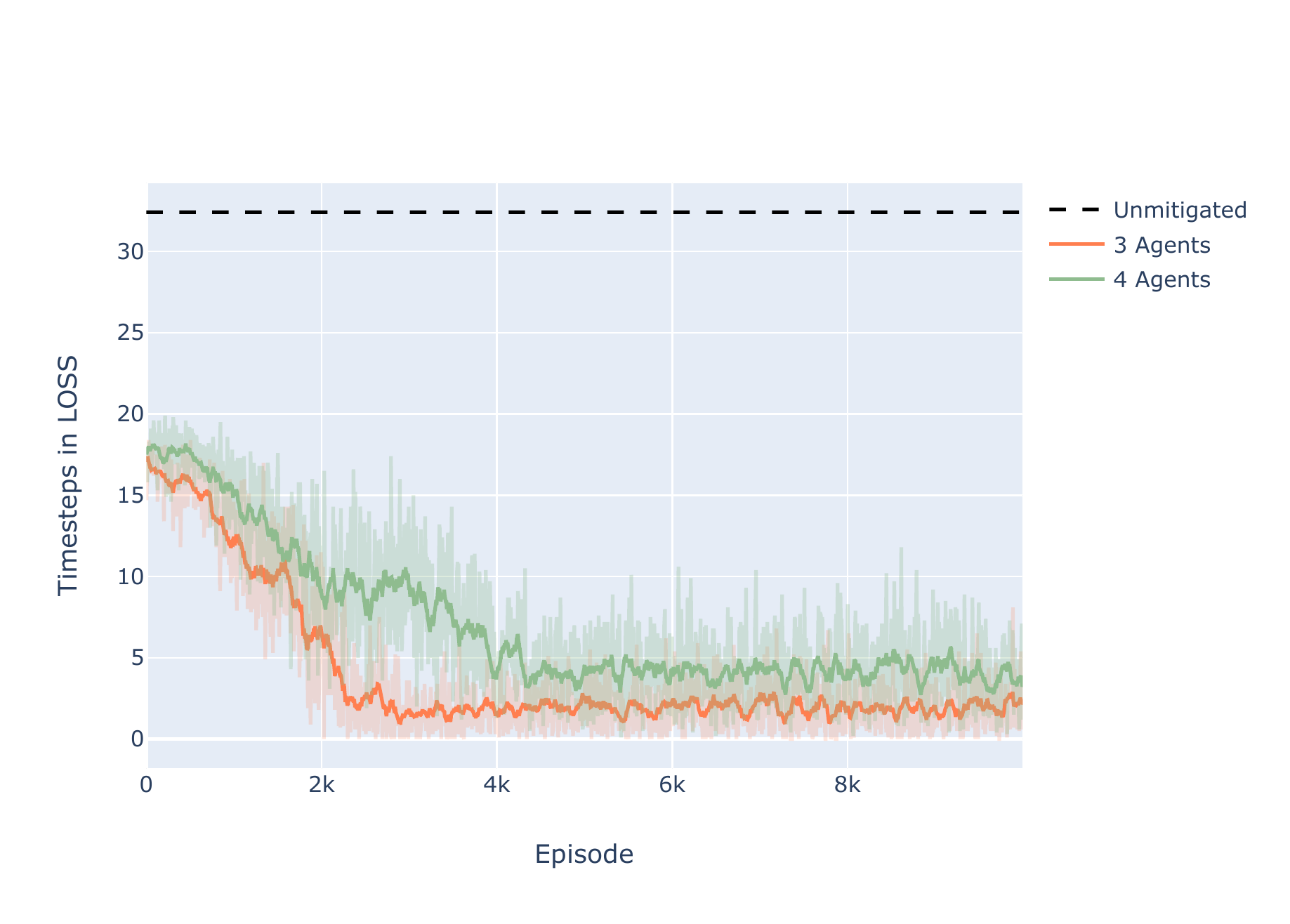}
\caption{Number of time steps spent in LOSS.}
\label{fig:tloss}
\end{figure}

\begin{table}[h]

\centering
\caption{Number of episodes compound conflicts solved}
\begin{tabular}{llllllll} 
\toprule
    &  & 3 Agents   &  & 4 Agents\\ 
\hline
 \\&  &  5650  &  & 4461  \\

\bottomrule
\end{tabular}
\label{noloss}
\end{table}

In addition to solving conflicts, it is desirable for agents not to spend too much in a LOSS, as this can increase the risk of collisions. Such information is shown in Figure \ref{fig:tloss}. The results shown there further confirm that the agents are able to improve their performance. In a similar trend, the case of the 3 agents compound conflict seems simpler to solve successfully, as the agents spend less than 5 seconds in a LOSS, with 5650 episodes not experiencing a LOSS (therefore no time steps in LOSS). 

\subsection{Agent behavior}
The results shown so far indicate that the agents manage to successfully learn how to solve the task. We not that the model converges fast and maintains its knowledge of the system, thus avoiding the common forgetting issues. 

\begin{figure}[h]
\centering
\includegraphics[width=\linewidth]{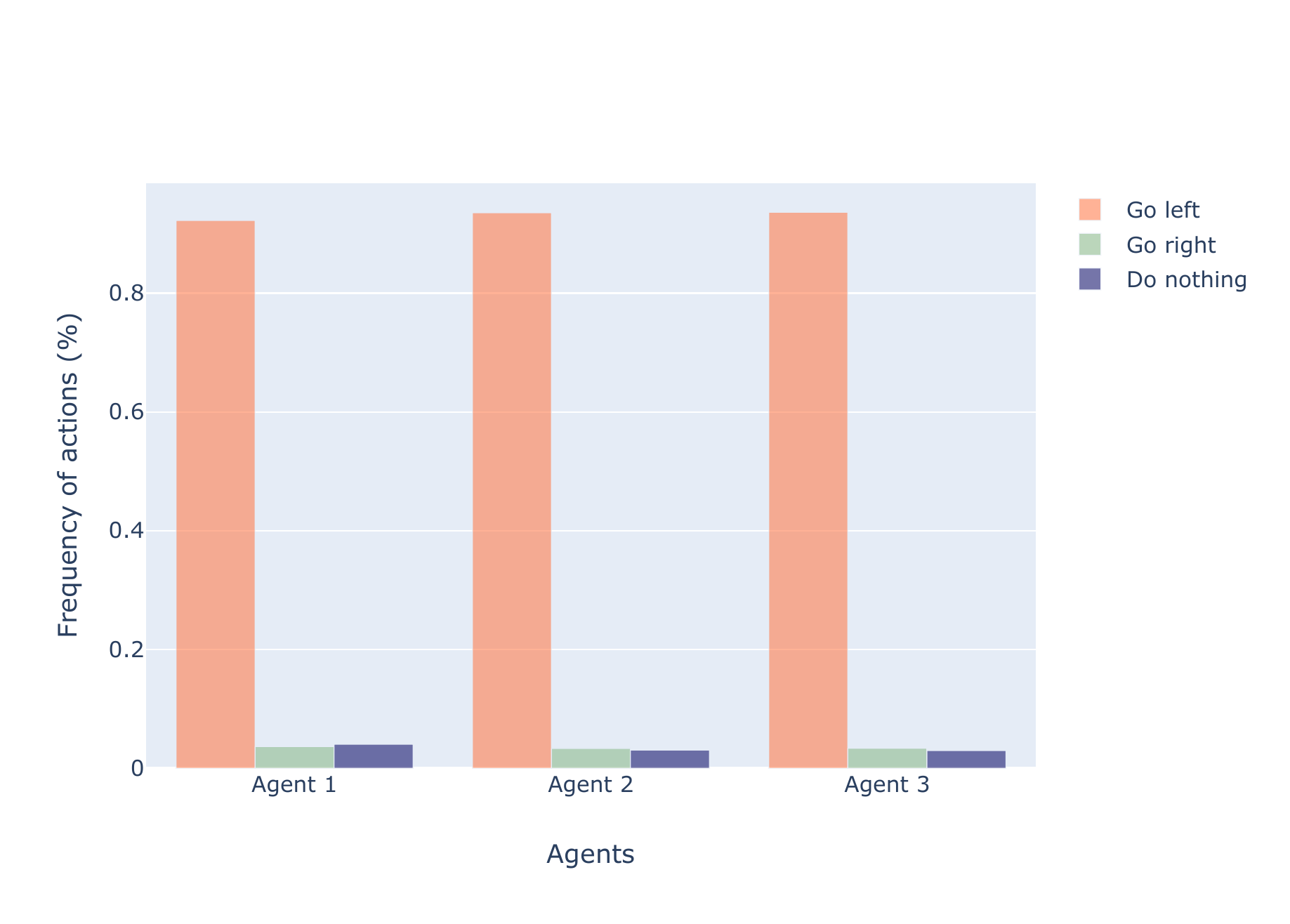}
\caption{Frequency of actions for the last 200 episodes for the compound conflict with 3 agents.}
\label{fig:act3}
\end{figure}

\begin{figure}[h]
\centering
\includegraphics[width=\linewidth]{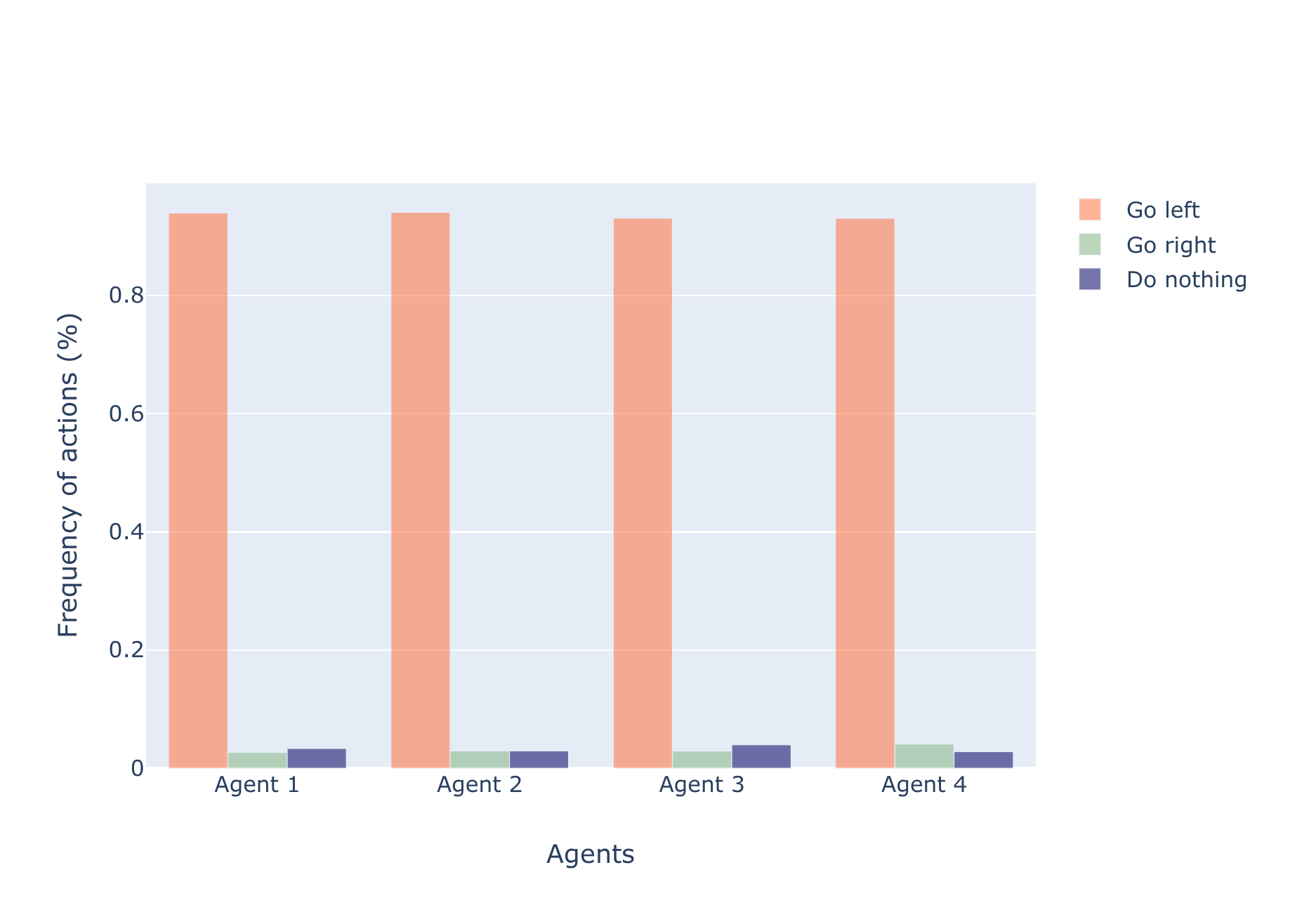}
\caption{Frequency of actions for the last 200 episodes for the compound conflict with 4 agents.}
\label{fig:act4}
\end{figure}

However, it is important to understand what strategies they have learned. This information is shown in Figures \ref{fig:act3} and \ref{fig:act4}. For the sake of simplicity, we only show the frequency of actions for the last 200 episodes. 

We note that in both settings, the agents take the \textit{go left} action in the majority of cases. While the direction of the action might not be as important, the learned strategy suggests that agents take the same action. This results in agents increasing the distance between them, as taking the same action head-on or crossing scenarios results in them going in different directions. However, in overtaking scenarios such a strategy does not immediately solve the conflict. Nevertheless, through the reward agents must learn that the conflict with the smallest CPA distance is the most urgent. As such, it can happen that agents prefer to delay the solution in an overtaking scenario, by taking several small changes in the same direction. While this is not immediately desirable, attempting to make a heading change to the opposite direction could create a more severe conflict with the head-on or crossing agents. 

In this work, we do not put any restrictions to the agents and do not inject expert knowledge in them, thus they start learning from a blank state. The results show that the agents are able to learn a strategy that successfully solves the compound conflicts in scenarios with 3 and 4 agents.

\section{Conclusions and Future Work}
In this paper, we tackle multi-UAV conflict resolution by modelling it as a MARL problem with cooperative agents. Air traffic is represented as a graph with aircraft as nodes. An edge is created between every two aircraft in a pairwise conflict. We use \textit{graph convolutional reinforcement learning}, which provides a communication mechanism between connected agents. This means that conflicting aircraft are allowed to communicate with each other and develop cooperative strategies. In order to formally define a multi-UAV conflict, we propose the concept of \textit{compound conflicts}, which are conflicts that have tight spatial and temporal boundaries.

We first train a model that learns how to solve compound conflicts with 3 agents. After that, the same model is retrained to to solve compound conflicts with 4 agents. As a result, we are able to refine the policies learned in the previous setting, while added agent learns a desirable policy.

Results show that the agents are able to improve their policies and thus solve the task. For both settings, we observe an improvement both in number of LOSS present and duration of LOSS with the majority of scenarios after convergence having no LOSS (i.e. the compound conflict is solved). Furthermore, the agents are able to discover a strategy that increases the overall distance between them. As such, they effectively learn to solve the most severe conflicts first and then solve the remaining conflicts while making sure that no new conflicts are created. 

However, there are several aspects that must be further researched. For instance, in this work we use a maximum of 4 agents in the scenario. In reality, the number of agents in a compound conflict can not be always decided beforehand, thus a solution that adapts to $\mathcal{N}$ agents must be sought. Furthermore, the reward function could be further elaborated to include terms that deal with the quality of solutions, such as optimizing for battery usage or number of actions taken. Finally, the action space can be extended to include solutions by speed or altitude changes.
\vspace{6pt} 



\section*{Acknowledgment}
This research has received funding in part from the SESAR Joint Undertaking under the European Union’s Horizon 2020 Research and Innovation Programme under grant agreement No 783287 and in part from Spanish National Project EU-TM No TRA2017-88724-R. The opinions expressed herein reflect the authors’ view only. Under no circumstances shall the SESAR Joint Undertaking be responsible for any use that may be made of the information contained herein.


%


\bibliographystyle{unsrtnat}
\bibliography{references}  

\begin{thebibliography}{46}
\providecommand{\natexlab}[1]{#1}
\providecommand{\url}[1]{\texttt{#1}}
\expandafter\ifx\csname urlstyle\endcsname\relax
  \providecommand{\doi}[1]{doi: #1}\else
  \providecommand{\doi}{doi: \begingroup \urlstyle{rm}\Url}\fi

\bibitem[{SESAR JU}(2016)]{SESARJU2016}
{SESAR JU}.
\newblock {European Drones Outlook Study}.
\newblock \emph{Single European Sky ATM Research}, \penalty0
  (November):\penalty0 93, 2016.
\newblock URL
  \url{http://www.sesarju.eu/sites/default/files/documents/reports/European_Drones_Outlook_Study_2016.pdf}.

\bibitem[Barrado et~al.(2020)Barrado, Boyero, Brucculeri, Ferrara, Hately,
  Hullah, Martin-Marrero, Pastor, Rushton, and
  Volkert]{Barrado2020U-spaceOperations}
Cristina Barrado, Mario Boyero, Luigi Brucculeri, Giancarlo Ferrara, Andrew
  Hately, Peter Hullah, David Martin-Marrero, Enric Pastor, Anthony~Peter
  Rushton, and Andreas Volkert.
\newblock {U-space concept of operations: A key enabler for opening airspace to
  emerging low-altitude operations}.
\newblock \emph{Aerospace}, 7\penalty0 (3):\penalty0 1--18, 2020.
\newblock ISSN 22264310.
\newblock \doi{10.3390/aerospace7030024}.

\bibitem[Prevot et~al.(2016)Prevot, Rios, Kopardekar, Robinson~III, Johnson,
  and Jung]{Prevot2016UASOperations}
Thomas Prevot, Joseph Rios, Parimal Kopardekar, John~E. Robinson~III, Marcus
  Johnson, and Jaewoo Jung.
\newblock {UAS Traffic Management (UTM) Concept of Operations to Safely Enable
  Low Altitude Flight Operations}.
\newblock \penalty0 (June):\penalty0 1--16, 2016.
\newblock \doi{10.2514/6.2016-3292}.

\bibitem[Zhang(2018)]{Zhang2018UOMSChina}
Jianping Zhang.
\newblock {UOMS in China}.
\newblock \emph{EU-China APP Drone Workshop}, pages 6--8, 2018.
\newblock URL
  \url{https://rpas-regulations.com/wp-content/uploads/2018/06/1.2-Day1_0910-1010_CAAC-SRI_Zhang-Jianping_UOMS-_EN.pdf}.

\bibitem[ica()]{icao}
{UTM - A Common Framework with Core Principles for Global Harmonization}.
\newblock
  \url{https://www.icao.int/safety/UA/Documents/UTM\%20Framework\%20Edition\%203.pdf}.
\newblock Accessed: 11.10.2021.

\bibitem[nas()]{nasa}
{NASA Conflict Management Model}.
\newblock
  \url{https://www.nasa.gov/sites/default/files/atoms/files/2020-johnson-nasa-faa.pdf}.
\newblock Accessed: 11.10.2021.

\bibitem[Radanovic et~al.(2019)Radanovic, Omeri, and Piera]{Radanovic_2019}
Marko Radanovic, Marsel Omeri, and Miquel~Angel Piera.
\newblock Test analysis of a scalable {UAV} conflict management framework.
\newblock \emph{Proceedings of the Institution of Mechanical Engineers, Part G:
  Journal of Aerospace Engineering}, 233\penalty0 (16):\penalty0 6076--6088,
  sep 2019.
\newblock \doi{10.1177/0954410019875241}.
\newblock URL \url{https://doi.org/10.1177%2F0954410019875241}.

\bibitem[Consiglio et~al.(2016)Consiglio, Muñoz, Hagen, Narkawicz, and
  Balachandran]{7778033}
Maria Consiglio, César Muñoz, George Hagen, Anthony Narkawicz, and Swee
  Balachandran.
\newblock Icarous: Integrated configurable algorithms for reliable operations
  of unmanned systems.
\newblock In \emph{2016 IEEE/AIAA 35th Digital Avionics Systems Conference
  (DASC)}, pages 1--5, 2016.
\newblock \doi{10.1109/DASC.2016.7778033}.

\bibitem[Johnson et~al.(2017{\natexlab{a}})Johnson, Petzen, and
  Tokotch]{Johnson2017Johnson_DetectAndAvoidHandout}
Sally~C Johnson, Alexander~N Petzen, and Dylan~S Tokotch.
\newblock {Johnson{\_}DetectAndAvoid Aviation 2017 Handout}.
\newblock \penalty0 (June), 2017{\natexlab{a}}.

\bibitem[Manfredi et~al.(2017)Manfredi, Jestin, Manfredi, Jestin, Introduction,
  Dasc, and Manfredi]{Manfredi2017}
Guido Manfredi, Yannick Jestin, Guido Manfredi, Yannick Jestin,
  An~Introduction, Ahead Dasc, and Guido Manfredi.
\newblock {An Introduction to ACAS Xu and the Challenges Ahead To cite this
  version : HAL Id : hal-01638049 An Introduction to ACAS Xu and the Challenges
  Ahead}.
\newblock 2017.

\bibitem[Cook et~al.(2017)Cook, Arnett, and Cohen]{Cook2017}
Brandon Cook, Tim Arnett, and Kelly Cohen.
\newblock {A Fuzzy Logic Approach for Separation Assurance and Collision
  Avoidance for Unmanned Aerial Systems}.
\newblock \emph{Modern Fuzzy Control Systems and Its Applications}, pages
  1--32, 2017.
\newblock \doi{10.5772/68126}.

\bibitem[Consiglio et~al.(2019)Consiglio, Duffy, Balachandran, Glaab, and
  Mu{\~{n}}oz]{Consiglio}
Maria Consiglio, Brendan Duffy, Swee Balachandran, Louis Glaab, and César
  Mu{\~{n}}oz.
\newblock {Sense and avoid characterization of the independent configurable
  architecture for reliable operations of unmanned systems}.
\newblock In \emph{13th USA/Europe Air Traffic Management Research and
  Development Seminar 2019}, 2019.

\bibitem[Mu{\~n}oz et~al.(2016)Mu{\~n}oz, Dutle, Narkawicz, and
  Upchurch]{Muoz2016UnmannedAS}
C{\'e}sar~A. Mu{\~n}oz, Aaron Dutle, Anthony Narkawicz, and Jason Upchurch.
\newblock Unmanned aircraft systems in the national airspace system: a formal
  methods perspective.
\newblock \emph{ACM SIGLOG News}, 3:\penalty0 67--76, 2016.

\bibitem[Weinert et~al.(2018)Weinert, Campbell, Vela, Schuldt, and
  Kurucar]{Weinert2018Well-clearRisk}
Andrew Weinert, Scot Campbell, Adan Vela, Dieter Schuldt, and Joel Kurucar.
\newblock {Well-clear recommendation for small unmanned aircraft systems based
  on unmitigated collision risk}.
\newblock \emph{Journal of Air Transportation}, 26\penalty0 (3):\penalty0
  113--122, 2018.
\newblock ISSN 23809450.
\newblock \doi{10.2514/1.D0091}.

\bibitem[McLain and Duffield(2017)]{McLain2017AWC}
Timothy~W. McLain and Matthew~O. Duffield.
\newblock A well clear recommendation for small uas in high-density,
  ads-b-enabled airspace.
\newblock 2017.

\bibitem[Modi et~al.(2016)Modi, Ishihara, Jung, Nikaido, D'souza, Hasseeb, and
  Johnson]{Modi2016ApplyingRN}
Hemil~C. Modi, Abraham~K. Ishihara, Jaewoo Jung, Ben~E. Nikaido, Sarah~N.
  D'souza, Hashmatullah Hasseeb, and Marcus Johnson.
\newblock Applying required navigation performance concept for traffic
  management of small unmanned aircraft systems.
\newblock 2016.

\bibitem[Jiang et~al.(2018)Jiang, Dun, Huang, and Lu]{jiang2018graph}
Jiechuan Jiang, Chen Dun, Tiejun Huang, and Zongqing Lu.
\newblock Graph convolutional reinforcement learning.
\newblock \emph{arXiv preprint arXiv:1810.09202}, 2018.

\bibitem[Skowron et~al.(2019)Skowron, Chmielowiec, Glowacka, Krupa, and
  Srebro]{skowron2019sense}
Michal Skowron, Witold Chmielowiec, Karolina Glowacka, Magdalena Krupa, and
  Adam Srebro.
\newblock Sense and avoid for small unmanned aircraft systems: Research on
  methods and best practices.
\newblock \emph{Proceedings of the Institution of Mechanical Engineers, Part G:
  Journal of Aerospace Engineering}, 233\penalty0 (16):\penalty0 6044--6062,
  2019.

\bibitem[Kuchar and Yang(2000)]{kuchar2000review}
James~K Kuchar and Lee~C Yang.
\newblock A review of conflict detection and resolution modeling methods.
\newblock \emph{IEEE Transactions on intelligent transportation systems},
  1\penalty0 (4):\penalty0 179--189, 2000.

\bibitem[Ribeiro et~al.(2020{\natexlab{a}})Ribeiro, Ellerbroek, and
  Hoekstra]{ribeiro2020review}
Marta Ribeiro, Joost Ellerbroek, and Jacco Hoekstra.
\newblock Review of conflict resolution methods for manned and unmanned
  aviation.
\newblock \emph{Aerospace}, 7\penalty0 (6):\penalty0 79, 2020{\natexlab{a}}.

\bibitem[Bertram and Wei(2020)]{bertram2020distributed}
Josh Bertram and Peng Wei.
\newblock Distributed computational guidance for high-density urban air
  mobility with cooperative and non-cooperative collision avoidance.
\newblock In \emph{AIAA Scitech 2020 Forum}, page 1371, 2020.

\bibitem[Yang and Wei(2020)]{yang2020scalable}
Xuxi Yang and Peng Wei.
\newblock Scalable multi-agent computational guidance with separation assurance
  for autonomous urban air mobility.
\newblock \emph{Journal of Guidance, Control, and Dynamics}, 43\penalty0
  (8):\penalty0 1473--1486, 2020.

\bibitem[Hu et~al.(2020)Hu, Yang, Wang, Wei, Ying, and Liu]{hu2020uas}
Jueming Hu, Xuxi Yang, Weichang Wang, Peng Wei, Lei Ying, and Yongming Liu.
\newblock Uas conflict resolution in continuous action space using deep
  reinforcement learning.
\newblock In \emph{AIAA AVIATION 2020 FORUM}, page 2909, 2020.

\bibitem[Ribeiro et~al.(2020{\natexlab{b}})Ribeiro, Ellerbroek, and
  Hoekstra]{ribeiro2020determining}
Marta Ribeiro, Joost Ellerbroek, and Jacco Hoekstra.
\newblock Determining optimal conflict avoidance manoeuvres at high densities
  with reinforcement learning.
\newblock \emph{10th SESAR Innovation Days}, 2020{\natexlab{b}}.

\bibitem[Isufaj et~al.(2021)Isufaj, Aranega~Sebastia, and
  Piera]{isufaj2021towards}
Ralvi Isufaj, David Aranega~Sebastia, and Miquel~Angel Piera.
\newblock Towards conflict resolution with deep multi-agent reinforcement
  learning.
\newblock In \emph{14th USA/Europe Air Traffic Management Research and
  Development Seminar (ATM2021)}, 2021.

\bibitem[Lowe et~al.(2017)Lowe, Wu, Tamar, Harb, Abbeel, and
  Mordatch]{lowe2017multi}
Ryan Lowe, Yi~Wu, Aviv Tamar, Jean Harb, Pieter Abbeel, and Igor Mordatch.
\newblock Multi-agent actor-critic for mixed cooperative-competitive
  environments.
\newblock \emph{arXiv preprint arXiv:1706.02275}, 2017.

\bibitem[Pham et~al.(2019)Pham, Tran, Alam, Duong, and
  Delahaye]{pham2019machine}
Duc-Thinh Pham, Ngoc~Phu Tran, Sameer Alam, Vu~Duong, and Daniel Delahaye.
\newblock A machine learning approach for conflict resolution in dense traffic
  scenarios with uncertainties.
\newblock In \emph{ATM Seminar 2019, 13th USA/Europe ATM R\&D Seminar}, 2019.

\bibitem[Brittain et~al.(2020)Brittain, Yang, and Wei]{brittain2020deep}
Marc Brittain, Xuxi Yang, and Peng Wei.
\newblock A deep multi-agent reinforcement learning approach to autonomous
  separation assurance.
\newblock \emph{arXiv preprint arXiv:2003.08353}, 2020.

\bibitem[Dalmau and Allard()]{dalmauair}
Ramon Dalmau and Eric Allard.
\newblock Air traffic control using message passing neural networks and
  multi-agent reinforcement learning.

\bibitem[Manfredi and Jestin(2016)]{manfredi2016introduction}
Guido Manfredi and Yannick Jestin.
\newblock An introduction to acas xu and the challenges ahead.
\newblock In \emph{2016 IEEE/AIAA 35th Digital Avionics Systems Conference
  (DASC)}, pages 1--9. IEEE, 2016.

\bibitem[Sutton et~al.(1998)Sutton, Barto, et~al.]{sutton1998introduction}
Richard~S Sutton, Andrew~G Barto, et~al.
\newblock \emph{Introduction to reinforcement learning}, volume 135.
\newblock MIT press Cambridge, 1998.

\bibitem[Bellman(1958)]{bellman1958dynamic}
Richard Bellman.
\newblock Dynamic programming and stochastic control processes.
\newblock \emph{Information and control}, 1\penalty0 (3):\penalty0 228--239,
  1958.

\bibitem[Fran{\c{c}}ois-Lavet et~al.(2018)Fran{\c{c}}ois-Lavet, Henderson,
  Islam, Bellemare, and Pineau]{franccois2018introduction}
Vincent Fran{\c{c}}ois-Lavet, Peter Henderson, Riashat Islam, Marc~G Bellemare,
  and Joelle Pineau.
\newblock An introduction to deep reinforcement learning.
\newblock \emph{arXiv preprint arXiv:1811.12560}, 2018.

\bibitem[Mnih et~al.(2013)Mnih, Kavukcuoglu, Silver, Graves, Antonoglou,
  Wierstra, and Riedmiller]{mnih2013playing}
Volodymyr Mnih, Koray Kavukcuoglu, David Silver, Alex Graves, Ioannis
  Antonoglou, Daan Wierstra, and Martin Riedmiller.
\newblock Playing atari with deep reinforcement learning.
\newblock \emph{arXiv preprint arXiv:1312.5602}, 2013.

\bibitem[Littman(1994)]{littman1994markov}
Michael~L Littman.
\newblock Markov games as a framework for multi-agent reinforcement learning.
\newblock In \emph{Machine learning proceedings 1994}, pages 157--163.
  Elsevier, 1994.

\bibitem[Wu et~al.(2020)Wu, Pan, Chen, Long, Zhang, and
  Philip]{wu2020comprehensive}
Zonghan Wu, Shirui Pan, Fengwen Chen, Guodong Long, Chengqi Zhang, and S~Yu
  Philip.
\newblock A comprehensive survey on graph neural networks.
\newblock \emph{IEEE transactions on neural networks and learning systems},
  32\penalty0 (1):\penalty0 4--24, 2020.

\bibitem[Veli{\v{c}}kovi{\'c} et~al.(2017)Veli{\v{c}}kovi{\'c}, Cucurull,
  Casanova, Romero, Lio, and Bengio]{velivckovic2017graph}
Petar Veli{\v{c}}kovi{\'c}, Guillem Cucurull, Arantxa Casanova, Adriana Romero,
  Pietro Lio, and Yoshua Bengio.
\newblock Graph attention networks.
\newblock \emph{arXiv preprint arXiv:1710.10903}, 2017.

\bibitem[Kipf and Welling(2016)]{kipf2016semi}
Thomas~N Kipf and Max Welling.
\newblock Semi-supervised classification with graph convolutional networks.
\newblock \emph{arXiv preprint arXiv:1609.02907}, 2016.

\bibitem[Gilmer et~al.(2017)Gilmer, Schoenholz, Riley, Vinyals, and
  Dahl]{gilmer2017neural}
Justin Gilmer, Samuel~S Schoenholz, Patrick~F Riley, Oriol Vinyals, and
  George~E Dahl.
\newblock Neural message passing for quantum chemistry.
\newblock In \emph{International conference on machine learning}, pages
  1263--1272. PMLR, 2017.

\bibitem[Zambaldi et~al.(2018)Zambaldi, Raposo, Santoro, Bapst, Li, Babuschkin,
  Tuyls, Reichert, Lillicrap, Lockhart, et~al.]{zambaldi2018relational}
Vinicius Zambaldi, David Raposo, Adam Santoro, Victor Bapst, Yujia Li, Igor
  Babuschkin, Karl Tuyls, David Reichert, Timothy Lillicrap, Edward Lockhart,
  et~al.
\newblock Relational deep reinforcement learning.
\newblock \emph{arXiv preprint arXiv:1806.01830}, 2018.

\bibitem[Koca et~al.()Koca, Isufaj, and Piera]{kocastrategies}
Thimjo Koca, Ralvi Isufaj, and Miquel~Angel Piera.
\newblock Strategies to mitigate tight spatial bounds between conflicts in
  dense traffic situations.

\bibitem[Hoekstra and Ellerbroek(2016)]{hoekstra2016bluesky}
Jacco~M Hoekstra and Joost Ellerbroek.
\newblock Bluesky atc simulator project: an open data and open source approach.
\newblock In \emph{Proceedings of the 7th International Conference on Research
  in Air Transportation}, volume 131, page 132. FAA/Eurocontrol USA/Europe,
  2016.

\bibitem[Shi et~al.(2020)Shi, Cai, Liu, and Yu]{inproceedings}
Ke~Shi, Kaiquan Cai, Zhaoxuan Liu, and Lanchenhui Yu.
\newblock A distributed conflict detection and resolution method for unmanned
  aircraft systems operation in integrated airspace.
\newblock pages 1--9, 10 2020.
\newblock \doi{10.1109/DASC50938.2020.9256477}.

\bibitem[Mullins et~al.(2013)Mullins, Holman, Foerster, Kaabouch, and
  Semke]{Mullins2013DynamicST}
Michael Mullins, Michael~W. Holman, Kyle Foerster, Naima Kaabouch, and William
  Semke.
\newblock Dynamic separation thresholds for a small airborne sense and avoid
  system.
\newblock 2013.

\bibitem[Ho et~al.(2018)Ho, Geraldes, Alves~Gonçalves, Cavazza, and
  Prendinger]{ICDR}
Florence Ho, Ruben Geraldes, Artur Alves~Gonçalves, Marc Cavazza, and Helmut
  Prendinger.
\newblock Improved conflict detection and resolution for service uavs in shared
  airspace.
\newblock \emph{IEEE Transactions on Vehicular Technology}, PP:\penalty0 1--1,
  12 2018.
\newblock \doi{10.1109/TVT.2018.2889459}.

\bibitem[Johnson et~al.(2017{\natexlab{b}})Johnson, Petzen, and
  Tokotch]{Johnson2017ExplorationOD}
Sally~C. Johnson, Alexander~N. Petzen, and Dylan~S. Tokotch.
\newblock Exploration of detect-and-avoid and well-clear requirements for small
  uas maneuvering in an urban environment.
\newblock 2017{\natexlab{b}}.

\end{thebibliography}

\end{document}